\documentclass[letterpaper]{article} % DO NOT CHANGE THIS
\usepackage{aaai24}  % DO NOT CHANGE THIS
\usepackage{times}  % DO NOT CHANGE THIS
\usepackage{helvet}  % DO NOT CHANGE THIS
\usepackage{courier}  % DO NOT CHANGE THIS
\usepackage[hyphens]{url}  % DO NOT CHANGE THIS
\usepackage{graphicx} % DO NOT CHANGE THIS
\usepackage{natbib}  % DO NOT CHANGE THIS AND DO NOT ADD ANY OPTIONS TO IT
\usepackage{caption} % DO NOT CHANGE THIS AND DO NOT ADD ANY OPTIONS TO IT
\usepackage{algorithm}
\usepackage{algorithmic}
\usepackage{multirow}
\usepackage{multicol}
\usepackage{amsfonts}
\usepackage{newfloat}
\usepackage{makecell}
\usepackage{listings}
\DeclareCaptionStyle{ruled}{labelfont=normalfont,labelsep=colon,strut=off} % DO NOT CHANGE THIS
\floatstyle{ruled}
\newfloat{listing}{tb}{lst}{}
\floatname{listing}{Listing}
\title{Self-Supervised Vision Transformers Are Efficient Segmentation Learners for Imperfect Labels}

\author {
    Seungho Lee\textsuperscript{\rm 1}\equalcontrib,
    Seoungyoon Kang\textsuperscript{\rm 1, \rm 2}\equalcontrib,
    Hyunjung Shim\textsuperscript{\rm 2}
}
\affiliations {
    \textsuperscript{\rm 1}Yonsei University, South Korea\\
    \textsuperscript{\rm 2}Korea Advanced Institute of Science and Technology, South Korea\\
    seungholee@yonsei.ac.kr, \{sy.kang, kateshim\}@kaist.ac.kr
}

\usepackage{bibentry}
\begin{document}

\maketitle

\begin{abstract}
This study demonstrates a cost-effective approach to semantic segmentation using self-supervised vision transformers (SSVT). By freezing the SSVT backbone and training a lightweight segmentation head, our approach effectively utilizes imperfect labels, thereby improving robustness to label imperfections. Empirical experiments show significant performance improvements over existing methods for various annotation types, including scribble, point-level, and image-level labels. The research highlights the effectiveness of self-supervised vision transformers in dealing with imperfect labels, providing a practical and efficient solution for semantic segmentation while reducing annotation costs. Through extensive experiments, we confirm that our method outperforms baseline models for all types of imperfect labels. Especially under the zero-shot vision-language-model-based label, our model exhibits 11.5\%p performance gain compared to the baseline.
\end{abstract}

\section{Introduction}
Semantic segmentation is a critical task in computer vision, involving the understanding of an image's semantics and the recognition of objects within it. Unlike image classification, this technique assigns a class to each pixel in an image in order to obtain dense predictions. Semantic segmentation is widely utilized in various fields that demand accurate and detailed predictions, such as autonomous driving and medical imaging~\cite{cordts2016cityscapes,dolz2018ivd}. Despite its importance, semantic segmentation faces challenges due to the requirement for highly precise pixel-level labels, which makes data preparation time-consuming and costly~\cite{cordts2016cityscapes}. This hinders the practical application of semantic segmentation.

To address the issue of the high annotation cost of semantic segmentation, there has been increasing interest in weakly supervised approaches that utilize annotations that are less expensive than pixel-level labels. These include methods ranging from scribble-level~\cite{pan2021scribble} to point-level~\cite{tel_liang2022tree,bearman2016s}, image-level~\cite{eps_lee2021railroad,SEAM_wang2020self}, and zero-shot approaches based on Vision-Language (VL) models~\cite{maskclip_zhou2022extract} (extracting templates from text). However, these cheaper labels used in weakly supervised approaches are imperfect compared to full supervision~\cite{adele_liu2022adaptive}: (1) they contain significant noise and (2) they provide far fewer labeled pixels. This imperfection can hinder the generalization ability of a model during training, which poses a limitation in real-world applications where low-cost model development is desired. Therefore, our focus is on effectively and efficiently utilizing these imperfect masks.

We propose a method that utilizes the shape prior of a self-supervised vision transformer (SSVT) to effectively leverage imperfect labels. Recent studies, such as DINO~\cite{dino_caron2021emerging,dinov2_oquab2023dinov2}, have demonstrated that when vision transformers are trained in a self-supervised manner, they develop features suitable for segmentation, including scene layout. To preserve the structural information inherent in SSVT, we propose using the SSVT as a backbone with a frozen state. We only train a lightweight segmentation head to assign classes based on shape-rich features. This approach maintains the robust shape prior of SSVT, resulting in features and segmentation results that are not biased toward imperfect data and exhibit a high level of generalization power. Furthermore, training only the segmentation head significantly reduces the number of parameter updates, thereby reducing the cost of training the entire model.

\begin{figure*}[t]
\begin{center}
\includegraphics[width=1.0\linewidth]{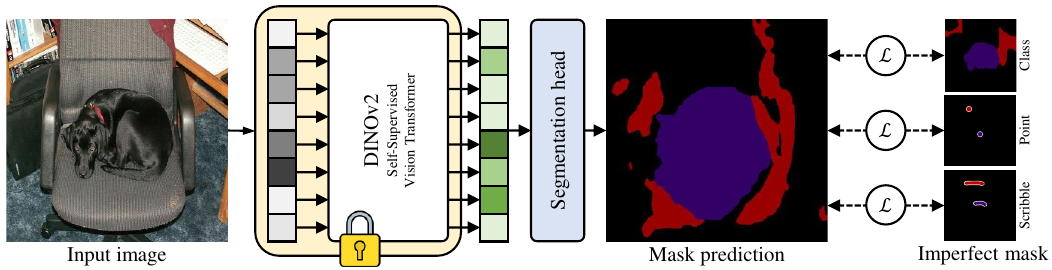}
\end{center}
\caption{Overview of our method. $\mathcal{L}$ represents the matching loss for each imperfect mask type (Equation~\ref{eqn:ssce}). For image-level label (class), $\mathcal{L}$ is pixel-wise cross-entropy. For others, $\mathcal{L}$ is masked pixel-wise cross-entropy. The backbone of the self-supervised vision transformer model is fixed during semantic segmentation training. Only the segmentation head is trained on imperfect masks and their corresponding images.}
\label{fig:overview}
\end{figure*}

Through empirical experiments, we validate that our method outperforms existing techniques across various types of weak annotations. Specifically, we observe performance improvements over the state-of-the-art TEL~\cite{tel_liang2022tree} for scribble and point level labels by 4.1\%p, and over ADELE~\cite{adele_liu2022adaptive} by 1.9\%p, which improves noisy image-level labels. Lastly, when using text-driven labels in a vision-language model like MaskCLIP~\cite{maskclip_zhou2022extract}, our method outperforms existing method~\cite{segformer_xie2021segformer} by 11.5\%p. We also confirm that various types of self-supervised vision transformers~\cite{ibot_zhou2021image,dino_caron2021emerging,dinov2_oquab2023dinov2} are effective in learning from imperfect labels compared to traditional methods for training segmentation networks.

In summary, our contributions are as follows:
\begin{itemize}
  \item We propose a cost effective strategy for training semantic segmentation network under imperfect labels, such as scribble, points, and noisy labels.
  \item We introduce a segmentation probe-centric training approach that effectively leverages the shape prior of SSVT models to enhance model generalizability.
  \item We validate the superiority of our model through various experiments across different types of imperfect labels.  
\end{itemize}

\section{Method}

We first introduce a brief training scheme for the semantic segmentation task using imperfect labels. For a given input image $x$ and its corresponding imperfect labels $y$, backbone network $f(\cdot)$ maps $x\in\mathbb{R}^{H\times W \times D}$ to its spatial feature $z\in\mathbb{R}^{H' \times W' \times Z}$. Then, the segmentation head $h(\cdot)$ evaluates the class probability of each pixel $\hat{y}\in\mathbb{R}^{H\times W \times C}$. We train the model using pixel-wise cross-entropy loss:
\begin{equation} \label{eqn:ssce}
    \mathcal{L} = - {{1}\over{HW}} \sum_{i=1}^{HW} [y_i \log \sigma(\hat{y_i})] \cdot M_i,
\end{equation}
where $\sigma(\cdot)$ is a sigmoid function. $M_i$ indicates the pixel mask according to scribble- and point-level labels. For image-level labels, $M_i$ is always set to 1. Unlike fully-supervised training, the label $y$ guiding the model training inevitably includes imperfections due to label acquisition methods. For scribble- and point-level labels, $y$ omits a large number of labels due to sparse annotation. Even worse, for image-level labels, $y$ is not accurate due to the inaccurate nature of pseudo-labeling methods, such as SEAM~\cite{SEAM_wang2020self} and EPS~\cite{eps_lee2021railroad}. Thus, it is crucial to build a robust model to handle imperfect label which includes both label insufficiency and noisy signals.

\begin{figure}[t]
\begin{center}
\includegraphics[width=1.0\linewidth]{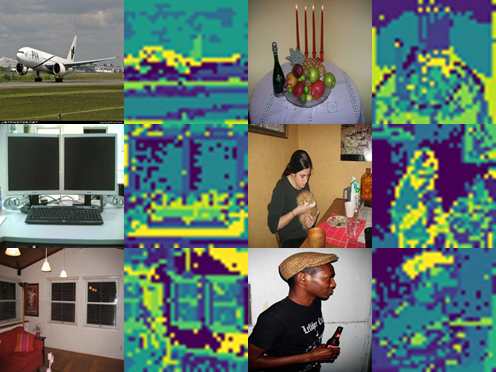}
\end{center}
\caption{DINOv2 feature analysis. For each image pair, the right image is the result of applying K-means clustering to each token from DINOv2 using the left image. Without any supervision, DINOv2 exhibits a strong shape prior, indicating that the objects are identifiable only with the K-means clustering.}
\label{fig:dinofeature}
\end{figure}

The recent discovery of DINO~\cite{dino_caron2021emerging} reveals that when training a vision transformer~\cite{vit_dosovitskiy2020image} in a self-supervised learning setting, it captures a scene layout suitable for segmentation in its features, a phenomenon that is not observed in traditional self-supervised learning methods. For example, we can observe the shape of an object in the DINOv2 feature~\cite{dinov2_oquab2023dinov2} in Figure~\ref{fig:dinofeature}. We propose an efficient and effective method for learning with imperfect labels, utilizing a pretrained self-supervised vision transformer (SSVT). Despite being trained in a self-supervised approach, the SSVT encapsulates significant shape characteristics of objects in images.

Specifically, we adopt the pretrained DINOv2~\cite{dinov2_oquab2023dinov2} model as our backbone. To effectively leverage the high-quality shape prior embedded in the SSVT, we employ the backbone in a fixed, non-updatable manner. We observed that when the SSVT with a high-quality shape prior is subjected to learning, it tends to fit the imperfect labels, thereby compromising its shape prior. Next, in order to assign class information to these high-quality features, we add and train a segmentation head. This segmentation head, consisting of a simple linear layer, transforms the patch tokens of the SSVT into predictions corresponding to the number of classes. Unlike the fixed backbone, the segmentation head is trained through gradient updates. This approach allows us to maintain the features of the SSVT, including the shape prior, while incorporating class information. This enables effective training even with imperfect data. Additionally, our proposed method is highly efficient because it only requires training the lightweight segmentation head, rather than the entire model, thanks to the fixed backbone.

\section{Experiments}
\subsection{Dataset}

We perform empirical analysis using the widely recognized benchmark dataset PASCAL VOC 2012~\cite{pascalvoc_everingham2010pascal}. This dataset consists of 21 categories (comprising 20 object types and one background category) and includes 1,464 training images, 1,449 validation images, and 1,456 test images. Consistent with standard practices in the field of semantic segmentation, we use an expanded augmented training set containing 10,582 images. For quantitative evaluation, we utilize the mean intersection-over-union (mIoU) metric to assess the accuracy of our segmentation models.

\subsection{Implementation Details}
We adopt ViT-B/14 from DINOv2 as the backbone network. We train the linear layer in the segmentation head and freeze the others. We use the SGD optimizer with a batch size of 10. The network is trained during 20K iterations with learning rate of 0.001. For data augmentation, we randomly apply cropping the input to $448 \times 448$, flipping, and color jittering.

\subsection{Comparison to SOTA}

\begin{table}[t!]
  \centering
  \small
  \begin{tabular}{l|lc|c}
  \hline
                                           & \multicolumn{2}{c|}{Baseline} & Ours                                      \\ \hline
Scribble                                   & TEL$^{'22}$             & 77.6       & 80.1                                      \\  \hline
Point                                      & TEL$^{'22}$             & 68         & 73.6                                      \\  \hline
\multicolumn{1}{l|}{\multirow{2}{*}{\makecell[l]{Class\\(Image-level)}}} & ADELE$^{'22}$           & 69.3       & \multicolumn{1}{c}{\multirow{2}{*}{71.2}} \\
                      & SegFormer$^{'21}$       & 65.6       & \multicolumn{1}{c}{}                      \\  \hline
Zero-shot VL                               & SegFormer$^{'21}$       & 26.9       & 38.4        \\  \hline \hline                             
\end{tabular}
    \caption{Quantitative evaluation of different types of imperfect label type. The cost of labeling decreases in the following order for each type of supervision: scribble, point, class (image-level), and zero-shot VL.}
    \label{tab:quanti-label-type}
\end{table}

\begin{figure}[t]
\begin{center}
\includegraphics[width=0.92\linewidth]{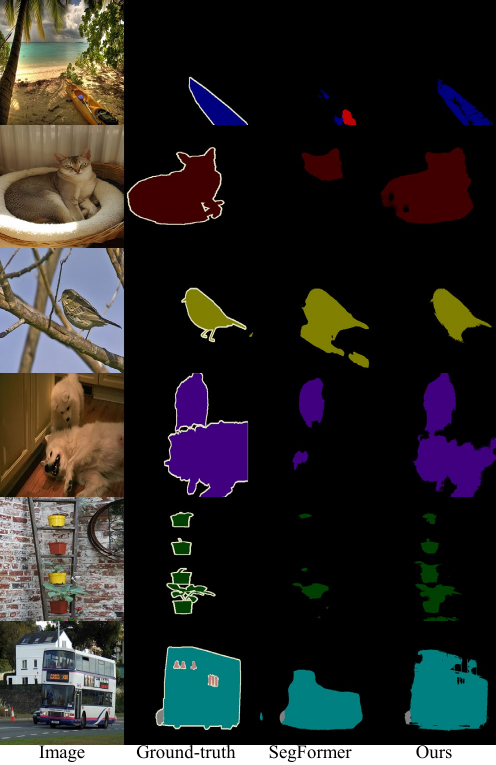}
\end{center}
\caption{Qualitative evaluation on image-level labels.}
\label{fig:qualitative}
\end{figure}

In this section, we demonstrate that our model can efficiently utilize various types of imperfect labels, resulting in high performance. Table~\ref{tab:quanti-label-type} compares the performance of our model with existing models based on the label acquisition cost. Firstly, compared to the state-of-the-art (SOTA) in scribble- and point-level label, TEL~\cite{tel_liang2022tree}, our model demonstrates an average performance improvement of 4.1\%p. Against one of the image-level label SOTAs, SegFormer~\cite{segformer_xie2021segformer}, our model exhibits a 5.6\%p performance increase. In Figure~\ref{fig:qualitative}, we consistently observe that our method produces a more precise segmentation mask than SegFormer. Especially in the last row, we observe that our method covers most of the boundary of the given image. Notably, even when compared with the ADELE~\cite{adele_liu2022adaptive}, which aims to address the over-confidence issue of models from imperfect image-level labels, our model still shows a 1.9\%p increase in performance. Additionally, in scenarios assuming extremely low labeling costs, where mask labels are derived from text via vision-language models, our model shows an 11.5\%p improvement over the baseline~\cite{segformer_xie2021segformer}. These results affirm that our proposed approach exhibits robustness across all types of imperfect data and achieves high performance.

\subsection{Ablation Study}
\subsubsection{Robustness analysis to imperfect label quality.}
\begin{table}[t!]
  \centering
  \small
    \begin{tabular}{l|c|cc}
    \hline
Method     & GT   & SEAM & EPS  \\ \hline
Pseudo-label      & -    & 63.6 & 69.4 \\ \hline
DeepLabV1  & 75.8 & 64.5 & 70.1 \\
DeepLabV3+ & 78.5 & 63.3 & 68.6 \\
SegFormer  & 82.8 & 65.5 & 69.0 \\
Ours       & 80.6 & 71.2 & 74.1 \\ \hline \hline
\end{tabular}
  
    \caption{Image-level label quality-based performance comparison. Quality indicates the mIoU between the pseudo-label of each method and the ground-truth. For each method, we evaluate mIoU along various types of pseudo-labels used for training the segmentation model.}
    \label{tab:quanti-sota}
\end{table}

We evaluate the robustness of our model according to the quality of the imperfect mask label. Table~\ref{tab:quanti-sota} presents the quantitative evaluation results for various qualities of image-level labels, comparing traditional weakly-supervised semantic segmentation (WSSS) methods and our method. A key observation is that the original WSSS models fail to significantly deviate from the quality of pseudo-labels, regardless of their model capacity (based on ground-truth, GT). When trained with GT, the performance of the models increases in this order: DeepLabV1~\cite{deeplabv1_liang2015semantic}, DeepLabV3+~\cite{deeplabv3_chen2018encoder}, SegFormer~\cite{segformer_xie2021segformer}. However, the performance difference becomes negligible when trained with pseudo-labels. Notably, in the case of the highest-quality pseudo-label, EPS~\cite{eps_lee2021railroad}, the model with the lowest performance, DeepLabV1, achieves a higher mIoU than the better-performing models. This suggests that when there are imperfections in pseudo-labels and insufficient data, models with more parameters, i.e., higher performance, are more likely to overfit on imperfect data.

In contrast, our method consistently demonstrates improved performance, indicating that our method is not dependent to the performance of pseudo-labels. This can be interpreted through the characteristics of SSVT. During the SSVT training process, the backbone learns superior features of the object, including structural information, primarily through view consistency. Since we use the pretrained SSVT as a frozen backbone, the high-quality features are not biased towards noisy imperfect labels, thus preventing the segmentation head from overfitting.

In conclusion, our experiments firstly demonstrate the vulnerability of the original WSSS to imperfect labels. Secondly, we observe that the stronger the backbone performance of the original WSSS, the greater the tendency for bias towards imperfect labels. Lastly, our approach confirms its potential as a robust model despite imperfect labels.

\subsubsection{Superiority of self-supervised vision transformer as a backbone.}
\begin{table}[t!]
  \centering
  \small
  \begin{tabular}{l|ll}
  \hline
Method     & SEAM & EPS  \\ \hline
Pseudo-label      & 63.6 & 69.4 \\ \hline
DINOv1     & 58.9 & 63.6 \\
ibot-L     & 65.2 & 70.0 \\
ibot-L/22k & 65.8 & 73.3 \\
DINOv2     & 71.2 & 74.1 \\ \hline \hline
\end{tabular}
    \caption{Self-supervised vision transformer performance across varying levels of imperfect label quality. All SSVT models are trained using our same strategy.}
    \label{tab:quanti-ssvt}
\end{table}

Table~\ref{tab:quanti-ssvt} demonstrates the quantitative performance of different SSVT models in relation to the quality of imperfect labels. We observe that as the performance of the SSVT model's backbone improves, its robustness to imperfect labels also increases. This outcome contrasts with the results of the original WSSS models, which tend to become contaminated with bias. From this experiment, it is evident that SSVT models consistently extract high-quality visual features, and the quality of these features determines their robustness to bias. Based on these experimental results, we select DINOv2, the most robust among SSVT models against imperfect labels, as our backbone.

\subsubsection{Source of robustness.}
\begin{table}[t!]
  \centering
  \small
  \begin{tabular}{cc|cc}
  \hline
\multirow{2}{*}{Method} & \multirow{2}{*}{Pretraining} & \multicolumn{2}{c}{Backbone strategy} \\ 
                        &                              & Freezing           & Tuning           \\ \hline
DeepLabV1               & Classification               & 64.6               & 64.5             \\
DeepLabV3+              & Classification               & 61.7               & 63.3             \\
SegFormer               & Classification               & 63.6               & 65.2             \\
DINOv2 (ours)                 & Self-supervised              & 71.2               & 64.5            \\ \hline \hline
\end{tabular}
    \caption{Performance analysis on backbone training strategies. Classification indicates model pretraining using ImageNet dataset~\cite{imagenet_deng2009imagenet}.}
    \label{tab:backbone-freeze}
\end{table}

Through previous experiments, we have confirmed the robustness of SSVT-based model against imperfect labels compared to original WSSS methods. We aim to analyze the source of this robustness, focusing on the backbone. Table~\ref{tab:backbone-freeze} presents the quantitative evaluation results for both SSVT-based and original WSSS methods using SEAM pseudo-labels. The evaluation considers scenarios where the backbone is either trained or not trained in conjunction with the segmentation head.

Observing the original WSSS first, we note that simply fixing the backbone (i.e., using ImageNet pretrained models) and training only the segmentation head does not necessarily result in improved performance. In contrast, our proposed model exhibits a drop in performance when the entire model is trained. However, this decline in performance remains at a level comparable to that of original WSSS. This suggests that even a backbone capable of extracting good features becomes biased toward imperfect labels during the fine-tuning stage when using target images and imperfect mask labels. From this experiment, we conclude that utilizing the high-capacity backbone obtained through the SSVT training process as-is is a crucial factor in situations with a scarcity of labels.

\subsection{Conclusion}

In this paper, we demonstrate an efficient approach for semantic segmentation using self-supervised vision transformers (SSVT) to handle imperfect labels. By leveraging the shape prior of SSVT, particularly the pretrained DINOv2 model, our method effectively overcomes the challenges of weakly supervised learning, where imperfect labeling induces bias to noisy signals. We maintain the backbone fixed during training to prevent overfitting to imperfect labels. We optimize a lightweight segmentation head for class assignment, which significantly reduces computational expenses.

Our experimental results show that our approach not only outperforms existing methods with various weak annotations but also demonstrates robustness against the bias of imperfect labels. This work contributes to the field by providing a cost-effective and efficient solution for weakly-supervised semantic segmentation, especially in situations where there is limited availability of accurate annotations. Our findings highlight the potential of self-supervised learning models in advancing practical applications across various domains.

\section{Acknowledgments} 
This work was supported by IITP grant funded by the Korea government(MSIT) and KEIT grant funded by the Korea government(MOTIE) (No. 2022-0-00680) and the Basic Science Research Program through the National Research Foundation of Korea (NRF) funded by the MSIP (NRF-2022R1A2C3011154) and the Ministry of Education (NRF-2022R1A6A3A13073319).

\bibliography{aaai24}

\end{document}